\documentclass[conference]{IEEEtran}
\IEEEoverridecommandlockouts

\usepackage{fontspec}
\usepackage{polyglossia}
\setdefaultlanguage{english}
\setotherlanguage{vietnamese}

\usepackage{stfloats}
\usepackage{tabularx}
\usepackage{booktabs}
\usepackage{caption}
\usepackage{cite}
\usepackage{hyperref}
\usepackage{amsmath,amssymb,amsfonts}
\usepackage{algorithmic}
\usepackage{graphicx}
\usepackage{float}     
\usepackage{textcomp}
\usepackage{xcolor}
\usepackage{caption}
\usepackage{subcaption}
\usepackage{booktabs}
\usepackage{tabularx}
\usepackage{array}
\usepackage{makecell}

\usepackage{tabularx}
\usepackage{makecell}
\usepackage{booktabs}
\usepackage{array}

\AtBeginDocument{
\selectlanguage{english}
\renewcommand{\tablename}{Table}
\renewcommand{\figurename}{Fig.}
\renewcommand{\appendixname}{Appendix}
}

\begin{document}

\title{ Complex-Text Robustness Evaluation and Failure Diagnosis for Low-Resource Multilingual Text-to-Speech  \\
}

\author{
\IEEEauthorblockN{
Tianlun Zuo$^{1}$, Ziyu Zhang$^{1}$, Tingzhi Mao$^{2}$, Zhonghua Fu$^{2}$, Lei Xie$^{1,*}$
}
\IEEEauthorblockA{
$^{1}$Audio, Speech and Language Processing Group (ASLP@NPU),\\
School of Computer Science, Northwestern Polytechnical University, Xi'an, China\\
$^{2}$iFLYTEK Company Ltd., Xi'an, China\\
tlzuo@mail.nwpu.edu.cn, lxie@nwpu.edu.cn
}
}

\maketitle

\begin{abstract}

% background and motivation
Low-resource multilingual text-to-speech (TTS) systems have expanded language coverage, but their robustness under complex text inputs remains insufficiently diagnosed.
Existing evaluations mainly focus on naturalness, speaker similarity, and content consistency using regular test sentences, while providing limited insight into how multilingual TTS systems fail when handling challenging inputs such as numbers, dates, named entities, long sentences, code-switched expressions, and punctuation-related structures.
% our work
This paper proposes a complex-text robustness diagnosis framework for low-resource multilingual TTS. We evaluate robustness from three dimensions: content consistency, language consistency, and generation stability. A multilingual robustness testing  scheme is designed for Thai, Vietnamese, Swahili, and Indonesian, covering ordinary sentences and multiple types of complex text inputs.
We further introduce automatic diagnostic metrics, including character error rate, language identification accuracy, and duration abnormal rate. To support input-level risk analysis before speech generation, we propose a lightweight Text Risk Score (TRS), which estimates synthesis risk from interpretable text features without manual annotation or model training.
% results
Experiments on three representative multilingual TTS systems, including OmniVoice, VoxCPM2, and MMS-TTS, show that complex text inputs expose systematic failure patterns that are not fully reflected by ordinary short-sentence evaluation. 
Different systems exhibit distinct vulnerabilities in number normalization, named entity handling, long-text generation, and code-switched input processing. Furthermore, TRS shows a positive correlation with content errors and duration abnormalities, demonstrating its usefulness as a low-cost pre-synthesis indicator for complex-text risk diagnosis in low-resource multilingual TTS.

\end{abstract}

\begingroup
\renewcommand\thefootnote{$^{*}$}
\footnotetext{Corresponding author.}
\endgroup


\section{INTRODUCTION}

% background
In recent years, text-to-speech (TTS) systems have gradually evolved from monolingual modeling for a few high-resource languages to multilingual speech generation with substantially broader language coverage. By leveraging multilingual training, cross-lingual transfer, speech representation learning, and unified generation frameworks, recent systems are able to synthesize speech for a much wider range of languages than earlier single-language TTS models \cite{intro_2021lowresource,intro_2024mms,intro_2024cosyvoice,intro_2024xtts}. This progress is particularly important for low-resource languages, where dedicated high-quality TTS systems are difficult to build due to limited paired speech data and linguistic resources.
However, broader language coverage does not necessarily guarantee reliable synthesis under complex real-world text inputs.

% existing systems
Existing low-resource and multilingual TTS studies mainly focus on improving synthesis quality under limited-resource conditions.
Early approaches commonly rely on multilingual training and cross-lingual transfer, where acoustic and linguistic representations learned from high-resource languages are transferred to low-resource target languages\cite{intro_2021lowresource, existing_2022laml}.
More recent systems further explore zero-shot or few-shot speech generation, speech prompting, semantic or acoustic token modeling, and unified multilingual generation frameworks, enabling TTS systems to synthesize speech across a wider range of languages and speakers\cite{intro_2024xtts, intro_2024cosyvoice, exsiting_2024valle}.
These studies have significantly improved the accessibility and scalability of low-resource TTS, shifting the focus from training a separate model for each language toward multilingual transfer and general-purpose speech generation.

% However
However, evaluating low-resource multilingual TTS only on ordinary sentences is insufficient for understanding its deployment reliability under complex text inputs. First, real TTS inputs often contain written expressions such as numbers, dates, named entities, long sentences, and punctuation-related structures, which place higher requirements on text normalization, pronunciation modeling, and prosody control\cite{exsiting_2015kestrel,exsiting_2019neuraltn, existing_2017semiotic}.
Second, these challenges are more pronounced in low-resource languages, where text front-end resources, pronunciation lexicons, G2P tools, and automatic evaluation systems are often less mature than those for high-resource languages\cite{existing_2016g2p, existing_2023nofrontend}.
Third, failures under complex text inputs are multidimensional. A model may generate natural-sounding speech while still producing content omissions, language inconsistencies, truncation, repetition, or abnormal duration. Recent robust TTS studies also show that language-model-based TTS systems can suffer from unstable prosody, high transcription error rates, and errors on hard sentences, suggesting that ordinary sentence evaluation is insufficient for deployment-oriented robustness diagnosis\cite{existing_2024valler, exsiting_2024ralle}
Therefore, a diagnosis-oriented evaluation is needed to reveal not only whether a multilingual TTS system can synthesize a language but also how and where it fails under complex text conditions.

% our work
To address this problem, this paper proposes a complex-text robustness diagnosis framework for low-resource multilingual TTS.
We evaluate robustness from three complementary dimensions: content consistency, language consistency, and generation stability.
Specifically, we design a multilingual complex-text robustness test covering Thai, Vietnamese, Swahili, and Indonesian, which represent different writing systems, linguistic characteristics, and resource conditions.
The test inputs include ordinary sentences, numeric and date expressions, named entities, long sentences, and punctuation-related expressions.
Based on this robustness test, we evaluate three representative multilingual TTS systems, including OmniVoice~\cite{intro_2026omnivoice}, VoxCPM2~\cite{intro_2025voxcpm}, and MMS-TTS~\cite{intro_2024mms}, using ASR-based character error rate, language identification accuracy, and duration abnormal rate. In addition, we introduce a lightweight \textbf{Text Risk Score} (TRS), which estimates potential synthesis risk from interpretable input-text features without manual annotation or additional model training.

Our main contributions are as follows:

\begin{itemize}

\item \textbf{Problem formulation and diagnostic framework.} We formulate complex-text robustness as an essential evaluation dimension for low-resource multilingual TTS, and propose a three-dimensional diagnostic framework covering content consistency, language consistency, and generation stability. This framework shifts the evaluation focus from "whether a system can synthesize a language" to "how and where it fails under complex real-world inputs."

\item \textbf{Multilingual complex-text robustness test and evaluation.} We construct a robustness test covering four linguistically diverse low-resource languages (Thai, Vietnamese, Swahili, and Indonesian) with six text categories. Using this test, we conduct a systematic robustness evaluation of three state-of-the-art multilingual TTS systems, revealing model- and language-specific failure patterns that are rarely observed in ordinary sentence evaluation.

\item \textbf{Text Risk Score (TRS) as a lightweight diagnosis tool.} We propose TRS, a training-free, interpretable metric that estimates input-level synthesis risk from textual features. We demonstrate that TRS is positively correlated with content and duration errors across systems, offering a low-cost indicator for pre-synthesis risk screening, while also clarifying its diagnostic scope by showing its limited predictive power for language identification errors.
\end{itemize}

Experimental results show that complex text inputs reveal failure patterns that are less visible in ordinary sentence evaluation. The observed errors vary across models, languages, and text categories, indicating that complex-text robustness is affected by both the modeling paradigm and language-specific properties. Further analysis shows that TRS is positively correlated with synthesis error indicators, suggesting its potential as a low-cost input risk diagnosis tool for low-resource multilingual TTS. These findings demonstrate that language coverage alone is not sufficient for evaluating low-resource multilingual TTS systems, and that complex-text robustness should be considered as an additional deployment-oriented evaluation dimension.

\section{RELATED WORK}

\subsection{Multilingual and Low-Resource Text-to-Speech}

Scaling TTS to low-resource languages is challenging because neural TTS systems usually require paired text-speech data and language-specific resources such as pronunciation lexicons, grapheme-to-phoneme models, and text front-end rules. Existing studies address this problem through multilingual training, cross-lingual transfer, and few-shot adaptation, where acoustic, linguistic, or phonetic knowledge learned from high-resource languages is transferred to low-resource target languages \cite{intro_2021lowresource,existing_2022laml,related_2022fewshot_phoneme}. Beyond adaptation, recent multilingual TTS systems further expand language coverage and zero-shot generation ability. MMS scales speech technology to more than 1,000 languages \cite{intro_2024mms}; YourTTS and XTTS explore multilingual zero-shot multi-speaker synthesis and cross-lingual voice cloning \cite{related_2022yourtts,intro_2024xtts}; and VALL-E, VALL-E X, and CosyVoice formulate TTS as speech-token or semantic-token generation for more flexible multilingual synthesis \cite{related_2023vallex, related_2024cosyvoice,exsiting_2024valle}.
These works show that multilingual modeling and transferable speech representations can improve low-resource TTS coverage and adaptation. However, most existing studies focus on language support, naturalness, speaker similarity, or ordinary sentence synthesis. Practical low-resource TTS still depends on text front ends, pronunciation representations, and language-specific resources, as reflected in recent low-resource TTS challenges such as Blizzard Challenge 2025 \cite{related_2025blizzard}. Therefore, while multilingual and low-resource TTS has made substantial progress in coverage and adaptation, its robustness under complex text inputs still needs explicit diagnosis.

\subsection{TTS Evaluation and Complex-Text Robustness}

TTS evaluation commonly combines subjective listening tests and automatic metrics. Mean Opinion Score (MOS) remains a widely used subjective measure for naturalness, while recent work further explores automatic MOS prediction to reduce the cost of large-scale human evaluation. MOSNet \cite{rw_2019mosnet} predicts human ratings from acoustic features for converted or synthesized speech, and VoiceMOS Challenge studies show that self-supervised speech representations can be effective for predicting MOS across diverse TTS and voice conversion systems \cite{rw_2022voicemos,rw_2022utmos}. More recent evaluation studies, such as TTSDS and TTSDS2, further argue that modern TTS evaluation should consider multiple factors, including prosody, speaker identity, and intelligibility, rather than relying on a single perceptual score \cite{rw_2024ttsds,rw_2025ttsds2}. These works improve scalable quality assessment, but they mainly evaluate generated speech after synthesis and do not explicitly diagnose which input texts are likely to trigger failures.

Complex text robustness is closely related to the text frontend of TTS. Before acoustic generation, written text often needs to be converted into spoken form, especially for non-standard words such as numbers, dates, abbreviations, symbols, and mixed-format expressions. Prior work on text normalization for speech applications shows that this conversion is non-trivial and strongly affects downstream speech synthesis quality \cite{rw_2017nsw,rw_2022wfst_lm_tn,rw_2022vietnamese_nsw}. 
In low-resource or multilingual settings, the problem becomes more difficult because normalization rules, pronunciation lexicons, grapheme-to-phoneme models, and language-specific preprocessing tools are often unavailable, expensive to build, or inconsistent across languages \cite{rw_2016g2p_almostany,rw_2023_no_frontend,rw_2024_lexicon_free_g2p}. As a result, complex written inputs may introduce errors before acoustic generation, including wrong verbalization, unstable pronunciation, or inappropriate prosodic boundaries. These front-end errors can further appear as omitted content, abnormal pauses, or unstable language control in synthesized speech \cite{rw_2022wfst_lm_tn,rw_2022_vietnamese_nsw,rw_2023_text_normalization_tts}.

Different from general TTS quality assessment, this paper focuses on diagnosis-oriented robustness evaluation. Instead of only asking whether the synthesized speech sounds natural, we examine whether a low-resource multilingual TTS system can preserve input content, maintain the target language, and generate stable audio under controlled complex text inputs. This perspective complements existing MOS-based and distribution-based evaluation methods by linking input-level text risk factors to output-level synthesis failures.

\section{Methodology}

\subsection{Overview}

\begin{figure*}[t]
    \centering
    \includegraphics[width=\textwidth]{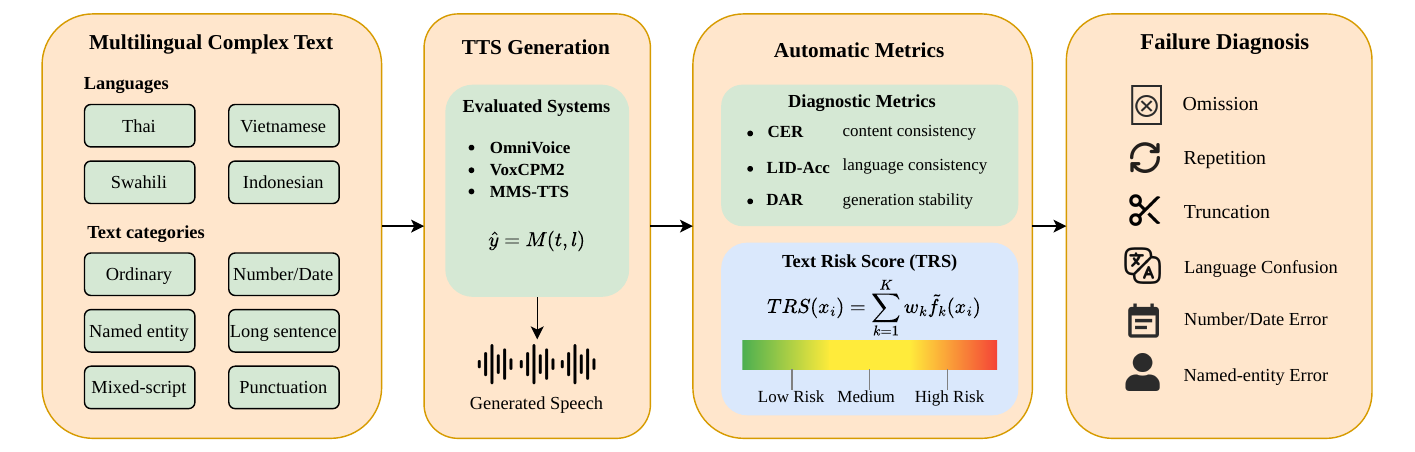}  % 或 0.9\textwidth
    \caption{Overview of the proposed complex-text robustness diagnosis framework.}
    \label{fig:overview}
\end{figure*}

This paper investigates complex-text robustness diagnosis for low-resource multilingual TTS. Unlike conventional TTS evaluation, which primarily focuses on naturalness, speaker similarity, or average quality on regular sentences, our objective is to diagnose deployment risks when TTS systems are exposed to complex and heterogeneous inputs.
Given a target language $l$, an input text $t$, and a TTS model $M$, the generated speech is denoted as
\begin{equation}
\hat{y}=M(t,l).
\end{equation}
We evaluate the reliability of $\hat{y}$ from three complementary dimensions: content consistency, language consistency, and generation stability. Content consistency examines whether the synthesized speech accurately preserves the textual content of the input. Language consistency examines whether the generated speech remains in the specified target language without language confusion. Generation stability examines whether the model produces valid, complete, and duration-consistent audio, rather than outputs with omissions, repetitions, truncation, or abnormal durations.
Based on these dimensions, we propose a diagnosis-oriented complex-text robustness evaluation framework, as illustrated in Fig.~\ref{fig:overview}. The framework consists of four stages.

First, in the complex-text robustness testing stage, we construct a multilingual evaluation set covering ordinary sentences and several high-risk categories: numbers and dates, named entities, long sentences, code-switched expressions, and punctuation-related constructions. These categories are designed to challenge text normalization, pronunciation modeling, language control, and long-form generation.

Second, in the speech generation stage, the same inputs are synthesized by three representative multilingual TTS systems: OmniVoice, VoxCPM2, and MMS-TTS, ensuring fair comparison under identical conditions.

Third, in the automatic diagnosis stage, we evaluate the generated speech using character error rate (CER) for content consistency, language identification accuracy (LID-Acc) for language consistency, and duration abnormal rate (DAR) for generation stability. These metrics provide complementary evidence for diagnosing different failure modes.

Finally, in the failure attribution stage, we combine text categories, automatic diagnostics, the proposed Text Risk Score (TRS), and manual sampling analysis to identify typical failure modes, including omissions, repetitions, truncation, language confusion, and verbalization errors. TRS provides a lightweight pre-synthesis risk indicator from interpretable text-level features, without requiring manual annotation, additional training, or access to generated speech.

Through this framework, we examine three aspects of complex-text robustness: the adequacy of ordinary sentence evaluation, model-specific failure patterns across languages and text categories, and the feasibility of using input text complexity as a prior risk indicator. The framework offers a low-cost, extensible diagnostic approach for pre-deployment evaluation, model selection, and text front-end optimization.

\subsection{Multilingual Complex Text robustness Set}

\subsubsection{Language Selection and Test Set Construction}
\label{sec:robustness_set}
To evaluate the robustness of low-resource multilingual TTS systems under complex text inputs, we select Thai, Vietnamese, Swahili, and Indonesian as the experimental languages. We select these languages as representative low-resource languages that cover different writing systems, linguistic characteristics, and regional distributions. This design enables us to analyze complex-text robustness under diverse language conditions while keeping the experimental scale controllable.

Specifically, Thai employs a non-Latin script and typically lacks explicit whitespace word boundaries in natural text, which may increase the difficulty of tokenization, prosodic boundary prediction, and front-end processing. Vietnamese utilizes an extended Latin alphabet with a rich system of diacritical tone marks, making it suitable for examining how TTS systems handle complex orthography and phonological distinctions. Swahili, a representative Bantu language written in the Latin script, has a relatively standardized orthography compared to many extremely low-resource languages; however, its speech corpora and pronunciation modeling tools remain substantially less developed than those of high-resource languages. Indonesian, also Latin-scripted, features highly regular spelling–pronunciation correspondences, making it a useful reference language with comparatively lower front-end complexity. Moreover, its frequent use in code-switched scenarios with regional languages (e.g., Javanese) allows us to assess model behavior on mixed-language inputs. By combining these four languages, we aim to systematically compare how differences in writing systems, linguistic properties, and resource availability collectively influence multilingual TTS robustness under complex text inputs.

For each language, we construct an experimental evaluation set following the same category taxonomy and construction rules. To ensure comparability across languages, the four languages share the same text categories, and we keep the number of samples and the length distribution of each category as consistent as possible. 
To facilitate replication and extension by future studies, we provide detailed category definitions, construction rules, statistical summaries, and representative examples.

\subsubsection{Text Categories and Construction Rules}

\begin{table*}[t]
\centering
\caption{Text categories, representative Vietnamese examples, and diagnostic targets in the complex-text robustness test.}
\label{tab:text_categories}
\small
\setlength{\tabcolsep}{4pt}
\renewcommand{\arraystretch}{1.18}
\begin{tabularx}{\textwidth}{
>{\raggedright\arraybackslash}p{0.17\textwidth}
>{\raggedright\arraybackslash}X
>{\raggedright\arraybackslash}p{0.23\textwidth}
>{\raggedright\arraybackslash}p{0.21\textwidth}}
\toprule
Category & Representative input & Main robustness factor & Diagnostic target \\
\midrule

\makecell[l]{Ordinary sentences}
& \emph{Nước trong ly vừa đủ lạnh.}
& Regular daily expressions
& Basic synthesis ability \\

\makecell[l]{Numbers and dates}
& \emph{Vui lòng đến khám vào ngày 12 tháng 1 năm 2026 lúc 09:30.}
& Digits, dates, and time expressions
& Text normalization and verbalization \\

\makecell[l]{Named entities}
& \emph{Tập đoàn Vingroup công bố kế hoạch phát triển mới.}
& Person, place, and organization names
& Entity pronunciation and OOV handling \\

\makecell[l]{Long sentences}
& \emph{Sau khi nhà trường thông báo thay đổi thời khóa biểu, nhiều phụ huynh phải điều chỉnh thời gian đưa đón con vào buổi chiều.}
& Long-context content preservation
& Omission, repetition, and truncation \\

\makecell[l]{Code-switched /\\mixed-script}
& \emph{Trang web hỗ trợ thanh toán bằng Visa và Mastercard.}
& Foreign words, abbreviations, and mixed scripts
& Language control and mixed-script handling \\

\makecell[l]{Punctuation-related}
& \emph{Cô giáo hỏi: "Ai muốn giải thích đáp án câu này?"}
& Questions, exclamations, quotes, and pauses
& Prosody, pauses, and punctuation handling \\

\bottomrule
\end{tabularx}
\end{table*}

We design six categories of test inputs to cover high-risk text patterns commonly encountered in practical low-resource TTS applications. Table~\ref{tab:text_categories} provides representative Vietnamese examples for readability. 
The same six-category taxonomy and construction rules are applied to all evaluated languages, including Thai, Vietnamese, Swahili, and Indonesian.
The first category is ordinary short sentences, which serve as a reference for basic synthesis ability. 
The second category is numbers and dates, which evaluates the handling of written forms such as numbers, dates, time expressions, prices, phone numbers, percentages, and measurement units.
The third category is named entities, including person names, place names, organization names, and international entities, which tests the robustness of entity pronunciation and out-of-vocabulary handling. 
The fourth category is long or compound sentences, which evaluates content preservation and generation stability under longer contexts. The fifth category is code-switched or mixed-script expressions, where English abbreviations, brand names, software names, or foreign entities are inserted into target-language sentences.
The sixth category is punctuation-related expressions, including questions, exclamations, pauses, quotation marks, colons, and parallel structures, which are used to examine whether the model can generate appropriate pauses and prosodic patterns.

The test texts are constructed according to category-specific rules. Ordinary short sentences are selected from daily expressions or public text resources and are controlled in length to provide relatively low-risk inputs. Numbers and dates are mainly generated using templates and then translated or localized into each target language. The templates include dates, time expressions, prices, phone numbers, temperatures, percentages, and measurement units. Named-entity texts contain both local entities in the target language and international entities, allowing us to compare how models handle local names and cross-lingual entities. Long-sentence inputs are controlled to be relatively long while still natural and grammatically plausible, avoiding artificially extreme or ill-formed cases. Code-switched or mixed-script inputs are constructed by inserting English abbreviations, application names, brand names, or international organization names into target-language sentences to simulate multilingual usage scenarios. Punctuation-related texts include interrogative, exclamatory, coordinated, paused, and quoted structures to test whether the model can produce reasonable pauses and intonation according to punctuation cues.

The construction process follows three principles. First, the texts should reflect realistic TTS inputs rather than artificially corrupted or adversarial examples. Second, the same category taxonomy and similar sample sizes are used across languages to ensure cross-language and cross-model comparability. Third, each category should correspond to a clear diagnostic target. For example, numbers and dates are mainly associated with text normalization risk, long sentences with content preservation and generation stability risk, and code-switched or mixed-script expressions with language control risk. This design allows the test set to function not only as an evaluation set, but also as a diagnostic tool for identifying where and why multilingual TTS systems fail under complex text inputs.

\subsubsection{Diagnostic Metrics}
\label{sec:metrics}
We design automatic diagnostic metrics from three dimensions: content consistency, language consistency, and generation stability. Unlike evaluation protocols that mainly rely on subjective naturalness scores, our focus is to examine whether complex text inputs lead to content loss, target-language deviation, or unstable audio generation. Therefore, we use character error rate (CER), language identification accuracy (LID-Acc), and duration abnormal rate (DAR) as the main automatic diagnostic metrics.

Character error rate (CER) measures content consistency between the input text and the synthesized speech. Specifically, we first transcribe the generated speech using a multilingual ASR system. The ASR transcription is then normalized and compared with the reference input text using character-level edit distance. CER is defined as:
\begin{equation}
    CER=\frac{S+D+I}{N},
\end{equation}
where $S$,$D$, and $I$ denote the numbers of substituted, deleted, and inserted characters, respectively, and $N$ denotes the number of characters in the reference text. Since word boundaries can be ambiguous or unstable in languages such as Thai, we use CER rather than word error rate (WER) as the main content consistency metric to reduce the influence of language-dependent tokenization tools.

Language identification accuracy (LID-Acc) measures whether the synthesized speech remains in the target language. For each successfully generated audio sample, we use a unified speech language identification model to predict its language label and compare it with the target language label. LID-Acc is defined as:
\begin{equation}
    LID\text{-}Acc=\frac{N_{\mathrm{correct}}}{N_{\mathrm{success}}},
\end{equation}
where $N_{\mathrm{correct}}$ denotes the number of samples whose predicted language label is consistent with the target language, and $N_{\mathrm{success}}$ denotes the number of successfully generated audio samples. This metric is mainly used to diagnose possible language confusion under low-resource or multilingual input conditions.

Duration abnormal rate (DAR) measures whether a model produces abnormal duration patterns, such as severe omission, repetition, truncation, or over-length generation. For each successfully generated sample, we compute the ratio between the audio duration and the length of the normalized evaluation reference:
\begin{equation}
r_i=\frac{\text{duration}_i}{|x^{\text{ref}}_i|},
\end{equation}
where $\mathrm{duration}_i$ denotes the duration of the $i$-th generated audio sample, $x^{\mathrm{ref}}_i$ denotes the evaluation reference used for CER computation. We use the normalized reference length rather than the raw input-text length because written forms and spoken forms can differ substantially, especially for numbers, dates, times, prices, phone numbers, and measurement units. This design avoids falsely treating number/date samples as duration-abnormal merely because their written inputs are much shorter than their spoken references.

Since speaking rate and duration distribution may vary across both languages and TTS systems, duration abnormality is identified within each model-language pair. Specifically, for each model-language pair, we compute the interquartile range (IQR) of the duration-to-reference-length ratio $r_i$. A sample is considered duration-abnormal if:
\[
r_i < Q_1 - 1.5 \times IQR \quad \text{or} \quad r_i > Q_3 + 1.5 \times IQR,
\]
where $Q_1$ and $Q_3$ are the first and third quartiles of the ratio distribution, and $IQR=Q_3-Q_1$. DAR is then defined as:
\begin{equation}
    DAR=\frac{N_{\mathrm{abnormal}}}{N_{\mathrm{success}}},
\end{equation}
where $N_{\mathrm{abnormal}}$ denotes the number of duration-abnormal samples and $N_{\mathrm{success}}$ denotes the number of successfully generated valid audio samples. This metric provides an automatic indication of generation instability, including possible omissions, repetitions, truncations, and abnormal duration expansion. It should be interpreted as a diagnostic indicator rather than a direct human-verified judgment, and representative abnormal cases are further inspected at the sample level.

\subsection{Text Risk Score}

\begin{table*}[t]
\centering
\caption{Risk features used in the proposed Text Risk Score.}
\label{tab:trs_features}
\small
\begin{tabular}{p{0.15\textwidth} p{0.10\textwidth} p{0.33\textwidth} p{0.34\textwidth}}
\hline
Feature & Symbol & Computation & Risk factor \\
\hline
Text length 
& $f_{\mathrm{len}}$ 
& Character or token count 
& Long-form omission, repetition, or truncation \\

Digit ratio 
& $f_{\mathrm{digit}}$ 
& Digit characters divided by text length 
& Number, date, price, and unit verbalization errors \\

Punctuation ratio 
& $f_{\mathrm{punc}}$ 
& Punctuation marks divided by text length 
& Pause, prosody, and syntactic boundary errors \\

Mixed-script ratio 
& $f_{\mathrm{mix}}$ 
& Non-target-script characters divided by text length 
& Language confusion and mixed-script handling errors \\

Named-entity flag 
& $f_{\mathrm{ent}}$ 
& Binary indicator of named-entity occurrence 
& Entity pronunciation errors and OOV-related shifts \\

Category prior 
& $f_{\mathrm{cat}}$ 
& Prior risk assigned by text category 
& Category-level risk from the robustness-test design \\
\hline
\end{tabular}
\end{table*}

To complement category-based robustness testing, we introduce a Text Risk Score (TRS) for estimating the potential synthesis risk of an input text before speech generation. 
Unlike trainable quality prediction models, TRS does not require manual annotations, additional model training, or access to the generated speech. 
Instead, it is computed only from interpretable text-level features. 
Therefore, TRS serves as a lightweight and low-cost input-level risk indicator for identifying texts that are more likely to trigger content errors, language confusion, or generation instability in low-resource multilingual TTS.

The design of TRS is based on the assumption that failures under complex text inputs are not entirely random, but are related to observable properties of the input text. 
Longer texts may increase the risk of omissions, repetitions, or truncation. 
Written expressions such as numbers, dates, prices, and units depend on text normalization and verbalization. 
Mixed-script expressions and foreign abbreviations may interfere with language control. 
Named entities and low-frequency words may lead to pronunciation errors or content shifts. 
Based on these considerations, we extract six rule-based risk features from each input text: text length, digit ratio, punctuation ratio, mixed-script ratio, named-entity flag, and category prior. 
These features are then normalized and combined into a single text-level risk score.

For an input text $x$, we define its raw risk feature vector as
\begin{equation}
\mathbf{f}(x) = [f_1(x), f_2(x), \ldots, f_K(x)],
\end{equation}
where $K$ denotes the number of risk features. 
The text length feature captures the risk of long-context generation. 
The digit ratio reflects the risk related to text normalization and verbalization. 
The punctuation ratio reflects potential difficulty in pause and prosody control. 
The mixed-script ratio measures the degree of cross-lingual or mixed-script input. 
The named-entity flag indicates whether the text contains person names, place names, organization names, brand names, or other entities. 
The category prior introduces the predefined risk level associated with the text category.
To reduce the influence of different writing systems and text-length distributions across languages, continuous features are normalized within each language using min-max normalization. 
Let $\mathcal{D}_l=\{x_i^{(l)}\}_{i=1}^{N_l}$ denote the set of input texts in language $l$, where $N_l$ is the number of texts in that language. 
For the $k$-th feature, the normalized feature value of text $x \in \mathcal{D}_l$ is computed as
\begin{equation}
\tilde{f}_k(x)=
\frac{
f_k(x)-\min_{x' \in \mathcal{D}_l} f_k(x')
}{
\max_{x' \in \mathcal{D}_l} f_k(x')-\min_{x' \in \mathcal{D}_l} f_k(x')+\epsilon
},
\end{equation}
where $\epsilon$ is a small constant used to avoid division by zero, $w_k$ denotes the weight of the $k$-th feature. In this work, we use equal weights for all features:
\begin{equation}
w_k=\frac{1}{K}, \quad k=1,2,\ldots,K.
\end{equation}
This setting avoids introducing additional training bias in the absence of manually annotated risk labels. It should be noted that TRS is not intended to precisely predict the synthesis error of each individual sample. Instead, it provides an interpretable text complexity measure for analyzing the relationship between input risk levels and TTS failure patterns.

The risk features used in TRS are summarized in Table~\ref{tab:trs_features}. Text length, digit ratio, punctuation ratio, and mixed-script ratio are directly computed from the input text. The named-entity flag is set to 1 if the text contains person names, place names, organization names, brand names, or international entities, and 0 otherwise. The category prior is assigned according to the predefined text category. Ordinary short sentences are assigned a lower prior risk, while numbers and dates, long sentences, and mixed-script expressions are assigned higher prior risks. Named entities and punctuation-related expressions are assigned medium prior risks. The category prior does not determine model performance by itself; rather, it incorporates the diagnostic prior of the robustness-test design into a unified risk score.

In the experimental analysis, we use TRS in two ways. First, to provide an intuitive comparison, we sort input texts by TRS within each language and divide them into three relative risk groups: low, medium, and high. This language-wise grouping avoids using a fixed global threshold across languages with different writing systems and text-length distributions. Specifically, for each language, texts are divided into three equal-sized groups according to the lower, middle, and upper tertiles of the TRS distribution. We then compare CER, LID-Acc, and DAR across the three risk groups. 
Second, to quantify the monotonic relationship between TRS and synthesis failures, we compute the Spearman correlation between TRS and each of the three diagnostic metrics: CER, DAR, and LID error rate. For CER, the correlation is defined as
\begin{equation}
\rho_s=\mathrm{Spearman}(TRS, CER).
\end{equation}
The same calculation is applied to DAR and LID error rate. A positive correlation indicates that higher-risk texts tend to produce higher synthesis errors, suggesting that TRS can serve as a pre-synthesis indicator of complex-text failure risk.

\section{Experiments} 

\subsection{Experimental Setup}
\label{sec:exp_setup}

We conduct experiments on four low-resource languages: Thai, Vietnamese, Swahili, and Indonesian. 
For each language, we construct 240 input texts, with 40 samples for each of the six text categories. This results in 960 input texts in total. Each evaluated TTS system synthesizes all 960 inputs, leading to 2880 synthesis tasks for the three systems. 
The detailed language selection criteria and category construction rules have been described in Section~\ref{sec:robustness_set}.
Our test texts were constructed following a category-specific protocol to enable fine-grained robustness diagnosis, a methodology aligned with recent multilingual TTS benchmarks MINT-Bench~\cite{testset_2026mint-bench}. 
Ordinary sentences and a portion of named entities were drawn from publicly available multilingual corpora, including FLEURS~\cite{testset_conneau2022fleursfewshotlearningevaluation} and Common Voice\cite{testset_ardila2020commonvoicemassivelymultilingualspeech}. The remaining high-risk categories were artificially constructed using category-specific templates and verified by native speakers to ensure linguistic naturalness and correctness.

To ensure a controlled comparison, all evaluated systems are tested under the same input-text, language, and category conditions. For each language, the same set of input texts is used for all models. Each input text is synthesized once by each model, and the generated audio is directly used for automatic diagnosis without repeated sampling or manual selection. No additional training data, model fine-tuning, speaker adaptation, voice cloning, or task-specific prompt engineering is applied. This setting is intended to reflect a practical off-the-shelf evaluation scenario, where released multilingual TTS systems are used without further customization.

For model inference, we follow the official inference pipeline of each system. 
When the model interface provides explicit language control, the target language is specified according to the corresponding model-supported language label. 
No model-specific text rewriting or manual correction is applied before synthesis, except for the standard input formatting required by the official implementation.

\subsection{Evaluated TTS Systems}
\label{sec:evaluated_systems}

We evaluate three representative multilingual TTS systems: OmniVoice, VoxCPM2, and MMS-TTS. 
They are selected because they support the evaluated languages, provide released checkpoints, and represent different multilingual TTS modeling routes. 
This selection enables a controlled comparison between recent speech generation model-style TTS systems and a broad-coverage multilingual neural TTS baseline under the same complex-text robustness-test setting.

\textbf{OmniVoice.} 
We evaluate OmniVoice using the released checkpoint \texttt{k2-fsa/OmniVoice}, which serves as a representative massive multilingual zero-shot TTS system in our experiments. 
It follows a speech foundation model-style generation route and emphasizes multilingual speech synthesis with acoustic-token-based generation. 
In this study, OmniVoice is used to examine how a recent large-scale multilingual zero-shot TTS system handles complex low-resource inputs, including numbers and dates, named entities, long sentences, mixed-script expressions, and punctuation-related expressions. 
Since the official release does not provide a clearly verified parameter scale in the information used in this work, we do not report its model size and only identify the released checkpoint used for evaluation.

\textbf{VoxCPM2.} 
We evaluate VoxCPM2 using the released checkpoint \texttt{openbmb/VoxCPM2}, which is selected as a representative tokenizer-free multilingual TTS system. 
According to its official release, VoxCPM2 is a 2B-parameter model trained on large-scale multilingual speech data and supports 30 languages. 
Different from speech-token-based systems, VoxCPM2 directly generates continuous speech representations without relying on an external discrete speech tokenizer. 
We include it to investigate whether this tokenizer-free multilingual generation route exhibits different robustness characteristics under complex text inputs, especially in terms of content preservation, language consistency, and duration stability.

\textbf{MMS-TTS.} 
We use the corresponding language-specific MMS-TTS checkpoints for the evaluated languages. 
MMS-TTS is selected as a broad-coverage multilingual neural TTS baseline rather than a language-model-based TTS system. 
It is part of the Massively Multilingual Speech project and provides TTS support for a large number of languages. 
In this work, MMS-TTS serves as a reference system for comparing broad language-coverage-oriented neural TTS with recent multilingual speech generation systems. 
This comparison helps analyze whether language coverage alone is sufficient for robust synthesis under complex real-world text inputs.

Overall, OmniVoice and VoxCPM2 represent recent multilingual speech generation model-style TTS routes, while MMS-TTS provides a broad-coverage neural TTS baseline. 
Evaluating these systems with the same languages, text categories, inference protocols, and automatic diagnostic metrics enables a controlled comparison of their failure patterns under complex input conditions.

\subsection{Evaluation Tools and Preprocessing}
\label{sec:eval_tools}

All generated audio files are first screened and then converted to a unified format before evaluation. 
Specifically, audio files are resampled to 16 kHz and converted to mono-channel waveforms. 
We record three types of invalid generation cases: missing waveform files, invalid audio files, and extremely short audio files with a duration shorter than 0.3 seconds. 
Invalid audio files refer to files that cannot be decoded correctly, contain empty waveforms, have zero duration after loading, or contain invalid waveform values such as NaN or Inf. 
These invalid outputs are excluded from ASR- and LID-based metric computation because reliable transcription or language identification cannot be performed on them. 
Their occurrence is retained for failure inspection. 
Unless otherwise specified, CER, LID-Acc, and DAR are computed over valid generated audio samples.

For content consistency evaluation, we use Whisper large-v3~\cite{exp_2022whisper} as a unified multilingual ASR backend. The language parameter is explicitly set to the target language when calling the ASR model, which disables automatic language detection and constrains the decoder to operate within the target-language vocabulary and acoustic space. This forced-target decoding ensures that the resulting transcription is always produced in the intended language space, preventing LID errors from contaminating CER measurements.  
The ASR model is run with deterministic beam-search decoding, with the beam size set to 5, and no random sampling is used during decoding. The resulting ASR transcription is then used to compute CER against the normalized reference text. 

For language consistency evaluation, we run Whisper-large-v3  without specifying the target language, allowing the backend to automatically infer the spoken language from the generated audio.
The detected language label is then compared with the target language label to compute LID-Acc. This decoupled design ensures that CER reflects genuine content preservation errors rather than LID-induced decoding bias, while LID-Acc independently quantifies the language identity of the synthesized speech as perceived by an automatic classifier.

For CER computation, we distinguish between the input text used for synthesis and the reference text used for evaluation. The original input text, denoted as $x_{\mathrm{raw}}$, is directly fed into the TTS system. The evaluation reference, denoted as $x_{\mathrm{ref}}$, is used for CER computation. For ordinary sentences, named entities, long sentences, mixed-script expressions, and punctuation-related expressions, $x_{\mathrm{ref}}$ is the original input text after basic normalization. For numbers and dates, $x_{\mathrm{ref}}$ is generated by converting written expressions into their expected spoken forms and then manually checked.

However, ASR systems may transcribe correctly spoken numerical expressions back into written digit forms. Directly comparing such ASR hypotheses with only spoken-form references may overestimate content errors. Therefore, for numbers-and-dates samples, we apply a format-robust reference matching strategy: CER is computed against both the spoken-form reference and the original written input, and the lower value is used as the final CER for reporting. For all other categories, CER is computed against $x_{\mathrm{ref}}$ only. This strategy reduces false penalties caused by ASR inverse text normalization while still preserving genuine errors such as incorrect digit reading, wrong date order, omissions, substitutions, and content shifts.

Before the edit-distance computation, the same normalization function $N(\cdot)$ is applied to both the evaluation reference and the ASR transcription. The normalization includes Unicode normalization, lowercasing where applicable, whitespace normalization, punctuation removal, and the standardization of basic symbols. Unless otherwise specified, CER in the following experiments refers to the final CER computed with this unified normalization and format-robust reference matching strategy.

For generation stability evaluation, duration abnormality is computed  in Section~\ref{sec:metrics}. 
In this computation, the text length is measured on the normalized reference text, i.e., $|\mathcal{N}(x_{\mathrm{ref}})|$, rather than on the raw input text. 
This choice avoids falsely treating number/date samples as over-length outputs due to a written-spoken format mismatch. 
DAR is then computed as the proportion of duration-abnormal samples among valid generated audio samples.

It should be noted that CER and LID-Acc are automatic diagnostic metrics rather than human-verified ground-truth judgments. 
Therefore, the same ASR backend, language detection backend, audio preprocessing procedure, decoding configuration, and duration-analysis procedure are applied to all systems and languages. 
This ensures consistent relative comparison across models, languages, and text categories, and the automatic results are further complemented by manual inspection of representative failure cases.

\section{Results and Analysis}

\subsection{Overall Robustness across Models and Languages}

We first report the overall robustness results across three TTS systems and four languages. Table~\ref{tab:overall_results} presents CER, LID-Acc, and DAR for each model-language pair. These metrics correspond to three complementary aspects of robustness: CER reflects content consistency, LID-Acc reflects whether the generated speech remains in the target language, and DAR reflects generation stability in terms of abnormal duration patterns. In this section, CER is computed using the unified text normalization and format-robust reference matching strategy described in the evaluation protocol. DAR is computed using the duration-to-normalized-reference-length ratio.

The results show that the three metrics reveal different types of robustness risks. OmniVoice achieves relatively low CER in most language settings, suggesting stronger content preservation under the current robustness-test conditions. However, its DAR is higher for Vietnamese and Swahili, indicating that low content error does not necessarily imply stable duration behavior. VoxCPM2 obtains the lowest CER on Indonesian, but its LID-Acc on Indonesian drops to 84.17\%, suggesting potential language-consistency issues. MMS-TTS shows a higher CER than OmniVoice in most language settings, especially for Thai and Vietnamese, while its DAR varies across languages. This indicates that MMS-TTS suffers from content realization errors and may also show abnormal duration behavior in some language settings.

Overall, these observations confirm that multilingual TTS robustness cannot be fully characterized by a single metric. A model may preserve textual content well but suffer from language inconsistency or abnormal duration, while another model may generate stable audio with weaker content accuracy. Therefore, the combination of CER, LID-Acc, and DAR provides a more complete diagnosis of low-resource multilingual TTS robustness.

\begin{table*}[t]
\centering
\caption{Overall robustness results across models and languages. CER is computed with unified text normalization and format-robust reference matching for number/date samples. DAR is computed using the normalized evaluation-reference length.}
\label{tab:overall_results}
\small
\begin{tabular}{llccc}
\hline
Model & Language & CER $\downarrow$ & LID-Acc $\uparrow$ & DAR $\downarrow$ \\
\hline
OmniVoice & Thai & 7.13\% & 100.00\% & 1.67\% \\
OmniVoice & Vietnamese & 2.09\% & 100.00\% & 17.50\% \\
OmniVoice & Swahili & 6.02\% & 98.32\% & 12.50\% \\
OmniVoice & Indonesian & 6.85\% & 97.48\% & 3.33\% \\
\hline
VoxCPM2 & Thai & 14.97\% & 100.00\% & 0.83\% \\
VoxCPM2 & Vietnamese & 29.39\% & 100.00\% & 4.17\% \\
VoxCPM2 & Swahili & 14.62\% & 96.67\% & 0.83\% \\
VoxCPM2 & Indonesian & 2.69\% & 84.17\% & 2.50\% \\
\hline
MMS-TTS & Thai & 22.04\% & 96.67\% & 0.83\% \\
MMS-TTS & Vietnamese & 13.81\% & 100.00\% & 5.83\% \\
MMS-TTS & Swahili & 10.36\% & 89.17\% & 5.83\% \\
MMS-TTS & Indonesian & 6.53\% & 91.67\% & 13.33\% \\
\hline
\end{tabular}
\end{table*}

\subsection{Effect of Complex Text Categories}

To examine whether complex text inputs expose additional risks beyond ordinary sentence synthesis, we compare the average robustness results across different text categories. As shown in Table~\ref{tab:category_results}, the results are averaged over all evaluated models and languages. The reported CER values follow the unified CER protocol, where number/date samples are evaluated with format-robust reference matching to reduce false penalties caused by ASR inverse text normalization.

The results show that numbers and dates remain the most challenging category in terms of content consistency. The CER of this category reaches 23.40\%, which is still higher than that of ordinary sentences, named entities, long sentences, and punctuation-related expressions. This indicates that even after reducing reference-format mismatch, numerical and date expressions continue to introduce substantial risks related to text normalization, verbalization, and content preservation.

Mixed-script expressions also produce a high CER of 17.77\%, suggesting that code-switched or mixed-script inputs remain difficult for multilingual TTS systems. This may be caused by foreign-word pronunciation, cross-lingual grapheme-to-phoneme conversion, or ASR transcription mismatch for acceptable transliterations. Named entities obtain a moderate CER of 6.78\%, reflecting the difficulty of entity pronunciation and out-of-vocabulary handling. In contrast, long sentences achieve the lowest CER of 2.06\%, indicating that sentence length alone is not necessarily the dominant source of content errors. If a long sentence does not contain high-risk tokens such as numbers, dates, foreign words, or named entities, it may still be synthesized reliably.

The LID-Acc and DAR results provide additional evidence beyond content consistency. Ordinary sentences show a relatively low LID-Acc of 94.17\%, which may be related to short utterance duration and limited language-specific acoustic cues. Numbers and dates obtain a high LID-Acc but also show the highest DAR of 17.92\%, suggesting that numerical expressions may affect generation stability or trigger repeated and abnormal-duration outputs in some systems. Punctuation-related expressions also show a higher DAR than named entities and mixed-script expressions, indicating that punctuation structures may affect pause control and duration distribution.

Representative diagnostic cases are provided in the Appendix. These cases further show that high CER values may arise from different sources, including genuine TTS content errors, number/date verbalization failures, ASR inverse text normalization, code-switched word confusion, named-entity substitution, and occasional repeated or hallucinated transcriptions.

To further interpret the category-level results, we conduct a sample-level inspection of representative cases from different models, languages, and text categories. Representative diagnostic cases are provided in the Appendix. These cases show that ASR-based CER may reflect different sources of errors, including genuine TTS content errors, number/date verbalization failures, ASR inverse text normalization, mixed-script transcription mismatch, and occasional repeated or hallucinated transcriptions. Therefore, category-level robustness should be interpreted from multiple perspectives rather than from CER alone.

\begin{table}[b]
\centering
\caption{Average robustness results across text categories. Results are averaged over all models and languages.}
\label{tab:category_results}
\small
\begin{tabular}{lccc}
\hline
Text category & CER $\downarrow$ & LID-Acc $\uparrow$ & DAR $\downarrow$ \\
\hline
Ordinary & 8.60\% & 94.17\% & 7.92\% \\
Num./Date & 23.40\% & 98.33\% & 17.92\% \\
Entity & 6.78\% & 97.50\% & 1.67\% \\
Long & 2.06\% & 97.92\% & 0.00\% \\
Mixed-script & 17.77\% & 97.92\% & 1.67\% \\
Punctuation & 9.65\% & 97.06\% & 5.42\% \\
\hline
\end{tabular}
\end{table}

\subsection{Model-specific Failure Patterns}

Table~\ref{tab:model_category_results} further compares model-specific failure patterns across text categories. To avoid an overly wide table, we focus on CER and DAR in this analysis. CER measures content-preservation errors, while DAR captures abnormal duration patterns such as omission, repetition, truncation, or over-length generation. LID-Acc is analyzed at the overall and language levels because language identification errors are more strongly related to language-specific acoustic similarity than to individual text categories.

For OmniVoice, the CER remains low on ordinary sentences, named entities, long sentences, mixed-script expressions, and punctuation-related inputs. Its CER on number/date inputs is 20.57\%, indicating that numerical and date expressions remain challenging even after format-robust correction. At the same time, OmniVoice shows a relatively high DAR on ordinary and number/date inputs, suggesting that this model may produce abnormal-duration outputs for some short or numerical samples.

For VoxCPM2, the number/date CER is 21.54\%, which is comparable to OmniVoice. However, VoxCPM2 shows a much higher CER on mixed-script expressions, reaching 32.09\%. This suggests that its main vulnerability in the current test set is not limited to numerical verbalization, but also includes code-switched and mixed-script input handling. Its DAR is generally lower than that of the other two systems, indicating that many VoxCPM2 failures are more related to content accuracy than duration abnormality.

MMS-TTS obtains the highest CER on several complex categories, including numbers and dates, named entities, and mixed-script expressions. Its DAR on number/date inputs reaches 23.75\%, showing that this category may also trigger duration abnormality in MMS-TTS. These results suggest that MMS-TTS tends to show weaker content accuracy under complex text inputs, while its duration stability is category-dependent, especially under numerical expressions.

Overall, the model-category analysis confirms that complex-text robustness should be examined by failure type rather than only by average model performance. A model may perform well on ordinary sentences but still fail on specific high-risk categories. Therefore, category-level diagnosis is necessary for practical low-resource multilingual TTS evaluation.

\begin{table*}[t]
\centering
\caption{Model-specific failure patterns across text categories. Results are averaged over four languages.}
\label{tab:model_category_results}
\small
\begin{tabular}{llcccccc}
\hline
Model & Metric & Ordinary & Num./Date & Entity & Long & Mixed-script & Punctuation \\
\hline
OmniVoice & CER $\downarrow$ & 3.04\% & 20.57\% & 2.13\% & 0.50\% & 2.90\% & 4.01\% \\
OmniVoice & DAR $\downarrow$ & 15.00\% & 28.75\% & 1.25\% & 0.00\% & 1.25\% & 6.25\% \\
\hline
VoxCPM2 & CER $\downarrow$ & 15.13\% & 21.54\% & 6.18\% & 0.79\% & 32.09\% & 16.79\% \\
VoxCPM2 & DAR $\downarrow$ & 2.50\% & 1.25\% & 1.25\% & 0.00\% & 2.50\% & 5.00\% \\
\hline
MMS-TTS & CER $\downarrow$ & 7.63\% & 28.08\% & 12.04\% & 4.88\% & 18.31\% & 8.16\% \\
MMS-TTS & DAR $\downarrow$ & 6.25\% & 23.75\% & 2.50\% & 0.00\% & 1.25\% & 5.00\% \\
\hline
\end{tabular}
\end{table*}

\subsection{Language-specific Robustness Analysis}

To connect the experimental results with language-specific properties, we summarize the average results for each language in Table~\ref{tab:language_results}. The results are averaged over the three evaluated TTS systems and all text categories.

Thai obtains the highest CER of 14.72\%, suggesting that it remains one of the most challenging languages in terms of content consistency. This may be related to its non-Latin writing system, lack of explicit whitespace word segmentation, and the difficulty of reliable ASR-based evaluation. Vietnamese obtains a CER of 15.10\% and a perfect LID-Acc of 100.00\%, indicating that language consistency does not necessarily guarantee content accuracy. In terms of generation stability, Vietnamese shows the highest DAR of 9.17\%, followed by Swahili and Indonesian, both at 6.39\%. Indonesian achieves the lowest CER of 5.36\%, but its LID-Acc is only 93.33\%, which may reflect either genuine language-control issues or limitations of the LID classifier when distinguishing Indonesian from acoustically or lexically similar languages.

These results indicate that language-specific robustness is affected by both linguistic properties and evaluation-tool reliability. Therefore, language-level conclusions should be interpreted together with category-level results and representative failure cases.

\begin{table}[t]
\centering
\caption{Language-specific robustness results. Results are averaged over three TTS systems and all text categories.}
\label{tab:language_results}
\small
\begin{tabular}{lccc}
\hline
Language & CER $\downarrow$ & LID-Acc $\uparrow$ & DAR $\downarrow$ \\
\hline
Thai & 14.72\% & 98.06\% & 1.11\% \\
Vietnamese & 15.10\% & 100.00\% & 9.17\% \\
Swahili & 10.33\% & 97.21\% & 6.39\% \\
Indonesian & 5.36\% & 93.33\% & 6.39\% \\
\hline
\end{tabular}
\end{table}

\subsection{Effectiveness of Text Risk Score}

We further evaluate whether the proposed Text Risk Score (TRS) can serve as a pre-synthesis indicator of complex-text failure risk. Since TRS is designed as a lightweight, text-level diagnostic score, it is not expected to precisely predict the error of each individual sample. Instead, we examine whether higher TRS values are associated with more severe synthesis failures at the group and correlation levels. We use CER, DAR, and LID-Err as diagnostic indicators. CER reflects content consistency risk, DAR reflects generation stability risk, and LID-Err reflects language consistency risk. LID-Err is defined as:
\begin{equation}
LID\text{-}Err = 1 - LID\text{-}Acc.
\end{equation}

For intuitive analysis, we divide input texts into low-, medium-, and high-risk groups by sorting TRS values within each language. Table~\ref{tab:trs_group_results} reports the average failure indicators under different TRS-based risk groups. The high-risk group obtains the highest CER, reaching 16.89\%, while the low-risk and medium-risk groups obtain 8.80\% and 9.50\%, respectively. This suggests that TRS captures part of the input-level complexity that affects content preservation.

The relationship between TRS and the other two indicators is weaker and should be interpreted cautiously. The high-risk group obtains a DAR of 11.43\%, which is higher than that of the medium-risk group, but duration abnormality is also model-dependent and can be affected by repeated or over-length outputs. LID-Err does not show a clear monotonic trend, indicating that language identification errors are not mainly determined by text-level complexity. Instead, they may be more strongly affected by acoustic realization, language similarity, utterance length, or limitations of the LID classifier.

\begin{table}[t]
\centering
\caption{Average failure indicators under different TRS-based risk groups. Results are averaged over all models and languages.}
\label{tab:trs_group_results}
\small
\begin{tabular}{lccc}
\hline
Risk group & CER $\downarrow$ & DAR $\downarrow$ & LID-Err $\downarrow$ \\
\hline
Low & 8.80\% & 5.76\% & 4.36\% \\
Medium & 9.50\% & 0.45\% & 1.35\% \\
High & 16.89\% & 11.43\% & 2.39\% \\
\hline
\end{tabular}
\end{table}

To further quantify the relationship between TRS and synthesis failures, we compute the Spearman rank correlation between TRS and each failure indicator. The correlation is first computed within each language and then averaged over the four languages. As shown in Table~\ref{tab:trs_correlation}, TRS shows a positive correlation with CER for all three systems, indicating that higher-risk texts tend to produce more severe content errors. The average TRS-CER correlation is 0.38, suggesting a moderate association between text-level risk and content-preservation difficulty.

The correlation between TRS and DAR is weaker and model-dependent, while TRS shows little positive correlation with LID-Err. This result clarifies the diagnostic scope of TRS: it is more suitable for identifying text-induced content risks than for predicting language identification errors. Therefore, TRS should be used as a coarse-grained pre-synthesis risk indicator rather than a substitute for ASR-based evaluation, LID analysis, or manual case inspection.

\begin{table}[t]
\centering
\caption{Spearman correlation between TRS and synthesis failure indicators. Each value is first computed within each language and then averaged over four languages.}
\label{tab:trs_correlation}
\small
\begin{tabular}{lccc}
\hline
Model & TRS-CER & TRS-DAR & TRS-LID-Err \\
\hline
OmniVoice & 0.42 & 0.43 & -0.01 \\
VoxCPM2 & 0.24 & 0.32 & -0.05 \\
MMS-TTS & 0.48 & -0.06 & -0.18 \\
\hline
Average & 0.38 & 0.23 & -0.08 \\
\hline
\end{tabular}
\end{table}

\section{Conclusion}

This paper investigates complex-text robustness in low-resource multilingual text-to-speech. Different from conventional evaluations that mainly focus on average naturalness or ordinary sentence synthesis, we propose a diagnosis-oriented framework that evaluates multilingual TTS systems from three dimensions: content consistency, language consistency, and generation stability. Based on this framework, we design a complex-text robustness test covering Thai, Vietnamese, Swahili, and Indonesian, and evaluate three representative multilingual TTS systems, including OmniVoice, VoxCPM2, and MMS-TTS. The test inputs include ordinary sentences, numbers and dates, named entities, long sentences, mixed-script expressions, and punctuation-related expressions.
Experimental results show that complex text inputs reveal failure patterns that are less visible in ordinary sentence evaluation. The observed errors vary across text categories, languages, and models, indicating that language coverage alone does not guarantee robustness under complex input conditions. In particular, numbers and dates, named entities, long sentences, and mixed-script inputs may lead to content errors, language inconsistency, or generation instability. Furthermore, the proposed Text Risk Score (TRS) provides a lightweight way to relate input-level text complexity to synthesis failures, supporting pre-synthesis risk analysis for low-resource multilingual TTS.
This work still has several limitations. First, the experiments cover only four languages and therefore cannot fully represent all low-resource languages or writing systems. Second, the evaluation mainly relies on automatic metrics and limited manual inspection, while fine-grained pronunciation accuracy and subjective naturalness are not fully analyzed. Third, TRS is based on rule-based text features and does not yet include a trainable risk prediction model. Future work will extend the language coverage and text categories, introduce native-speaker evaluation and more fine-grained pronunciation analysis, and explore text-front-end correction, risk-aware decoding, and synthesis-result filtering to improve the robustness of low-resource multilingual TTS in real-world applications.

\section*{Appendix: Representative Diagnostic Cases}
\label{app:case_analysis}

To better understand the category-level CER trends, we manually inspect representative sample-level cases from different models, languages, and text categories. Table~\ref{tab:appendix_case_analysis} summarizes typical cases. The purpose of this analysis is not to replace the quantitative metrics, but to clarify the sources of high CER values. In particular, high CER may be caused by genuine TTS content errors, number/date verbalization failures, mismatch between spoken-form references and ASR inverse text normalization, code-switched word confusion, named-entity substitution, or occasional repeated and hallucinated transcriptions.

\begin{table*}[t]
\centering
\caption{Representative diagnostic cases for interpreting category-level CER results. CER follows the final evaluation protocol used in the main experiments. For number/date samples, format-robust reference matching is applied to reduce false penalties caused by ASR inverse text normalization.}
\label{tab:appendix_case_analysis}
\scriptsize
\setlength{\tabcolsep}{3pt}
\renewcommand{\arraystretch}{1.15}
\begin{tabularx}{\textwidth}{
>{\raggedright\arraybackslash}p{0.08\textwidth}
>{\raggedright\arraybackslash}p{0.10\textwidth}
>{\raggedright\arraybackslash}p{0.20\textwidth}
>{\raggedright\arraybackslash}p{0.22\textwidth}
>{\raggedright\arraybackslash}p{0.22\textwidth}
>{\raggedright\arraybackslash}p{0.12\textwidth}
}
\hline
Category & Model / Lang. & TTS input & Spoken reference & ASR transcription & Diagnosis \\
\hline

Num./Date
& OmniVoice / Vi.
& \makecell[l]{Vui lòng đến khám vào ngày\\12 tháng 1 năm 2026 lúc 09:30.}
& \makecell[l]{Vui lòng đến khám vào ngày\\mười hai tháng một năm hai nghìn\\hai mươi sáu lúc chín giờ ba mươi.}
& \makecell[l]{Vui lòng đến khám vào ngày\\12 tháng 1 năm 2026, lúc 09.30.}
& \makecell[l]{ASR inverse text\\normalization;\\CER: 0.00\%} \\

\hline

Num./Date
& VoxCPM2 / Id.
& \makecell[l]{Nomor telepon toko ini adalah\\021-3456-7890.}
& \makecell[l]{Nomor telepon toko ini adalah\\nol dua satu tiga empat lima enam\\tujuh delapan sembilan nol.}
& \makecell[l]{Nomor telepon toko ini adalah\\021-3456-7890.}
& \makecell[l]{ASR inverse text\\normalization;\\CER: 0.00\%} \\

\hline

Num./Date
& MMS-TTS / Vi.
& \makecell[l]{Phòng họp ở tầng 15,\\phòng 1508.}
& \makecell[l]{Phòng họp ở tầng mười lăm,\\phòng một năm không tám.}
& \makecell[l]{Phong hợp nữ Tân Phong.}
& \makecell[l]{Number content loss;\\CER: 54.17\%} \\

\hline

Num./Date
& VoxCPM2 / Sw.
& \makecell[l]{Kasi ya wastani ya intaneti\\ni 300 Mbps.}
& \makecell[l]{Kasi ya wastani ya intaneti ni\\megabiti mia tatu kwa sekunde.}
& \makecell[l]{Kwa hivyo, kwa hivyo,\\kwa hivyo, kwa hivyo, ...}
& \makecell[l]{Repetition output;\\insertion-dominated\\error;\\Uncapped CER:\\634.00\%} \\

\hline

Code-switched
& MMS-TTS / Vi.
& \makecell[l]{Chúng tôi sẽ gửi newsletter\\mới vào thứ Sáu.}
& \makecell[l]{Chúng tôi sẽ gửi newsletter\\mới vào thứ Sáu.}
& \makecell[l]{Chúng tôi sẽ đợi next time\\mới vào thứ sáu.}
& \makecell[l]{Code-switched\\English word\\confusion;\\CER: 25.71\%} \\

\hline

Named entity
& MMS-TTS / Vi.
& \makecell[l]{Elon Musk nói về dự án\\Starlink ở Đông Nam Á.}
& \makecell[l]{Elon Musk nói về dự án\\Starlink ở Đông Nam Á.}
& \makecell[l]{Trump nói về dự án Stalin\\ở Đông Nam à?}
& \makecell[l]{Entity substitution;\\CER: 29.41\%} \\

\hline

Long
& MMS-TTS / Id.
& \makecell[l]{Jika mahasiswa mengumpulkan\\dokumen minggu ini, dosen dapat\\memeriksa dan memberi masukan\\sebelum hari presentasi.}
& \makecell[l]{Same as TTS input.}
& \makecell[l]{Jika mahasiswa mengumpulkan\\dokumen minggu ini, dosen dapat\\memeriksa dan memberi masukan\\sebelum hari presentasi.}
& \makecell[l]{Long but clean input;\\CER: 0.00\%} \\

\hline
\end{tabularx}
\end{table*}

The first two number/date cases show that CER may be overestimated if only the spoken-form reference is considered. In these cases, the synthesized content is largely consistent with the input, but the ASR system transcribes the spoken numbers back into written digit forms. This observation motivates the format-robust reference matching strategy used for number/date samples in the main experiments.

However, the high CER of number/date inputs cannot be attributed only to reference-format mismatch. The MMS-TTS Vietnamese example shows a genuine content realization failure, where key numerical information is not preserved in the ASR transcription. The Swahili example further shows an extreme repeated-output case. Since CER is not upper-bounded when insertion errors dominate, repeated or hallucinated transcriptions may produce CER values greater than 100\%. Such samples should be interpreted as outliers related to generation instability or ASR transcription behavior, rather than as ordinary pronunciation errors.

The code-switched and named-entity examples show different failure mechanisms. Code-switched words may be confused with acoustically similar English expressions, as shown by the substitution of ``newsletter'' with ``next time''. Named entities may also be replaced by acoustically or semantically unrelated words, as shown by the confusion between ``Elon Musk'' and ``Trump'' and between ``Starlink'' and ``Stalin''. In contrast, the long-sentence example shows that long inputs can still be synthesized accurately when they do not contain high-risk tokens. This supports the conclusion that sentence length alone is not the dominant factor in complex-text robustness; rather, specific written patterns such as numbers, code-switched words, and named entities play a more decisive role.

\bibliographystyle{IEEEbib}
\bibliography{arxiv}

\vspace{12pt}

\end{document}